\documentclass[final,5p,times,twocolumn]{elsarticle}
\usepackage{picins}
\usepackage{amsfonts}
\usepackage{graphicx}

\usepackage{amssymb}

\journal{Neurocomputing}
\begin{document}
\begin{frontmatter}

\title{Optimized Projection for Sparse Representation Based Classification}

\author[tongji,CAS,USTC]{Can-Yi Lu}
\ead{canyilu@mail.ustc.edu.cn}
\author[tongji]{De-Shuang Huang\corref{cor}}
\ead{huangdeshuang@yahoo.com}
\address[tongji]{School of Electronics and Information Engineering, Tongji University 4800 Caoan Road, Shanghai 201804, China}
\address[CAS]{Hefei Institute of Intelligent Machines, Chinese Academy of Sciences, Hefei, China}
\address[USTC]{Department of Automation, University of Science and Technology of China, Hefei, China}
\cortext[cor]{Corresponding author.}
\begin{abstract}
Dimensionality reduction (DR) methods have been commonly used as a principled way to understand the high-dimensional data such as facial images. In this paper, we propose a new supervised DR method called Optimized Projection for Sparse Representation based Classification (OP-SRC), which is based on the recent face recognition method, Sparse Representation based Classification (SRC). SRC seeks a sparse linear combination on all the training data for a given query image, and make the decision by the minimal reconstruction residual. OP-SRC is designed on the decision rule of SRC, it aims to reduce the within-class reconstruction residual and simultaneously increase the between-class reconstruction residual on the training data. The projections are optimized and match well with the mechanism of SRC. Therefore, SRC performs well in the OP-SRC transformed space. The feasibility and effectiveness of the proposed method is verified on the Yale, ORL and UMIST databases with promising results.
\end{abstract}
\end{frontmatter}
\section{Introduction}
In many application domains, such as appearance-based object recognition, information retrieval and text categorization, the data are usually provided in high-dimensional form. One of the problems is the so-called "curse of dimensionality" \cite{PRreview}, which is a well known but not entirely well-understood phenomenon. Limited data lie in high-dimensional space, and important features are not so much. Moreover, it has been observed that a large number of features may actually degrade the performance of classifiers if the number of training samples is small relative to the number of features \cite{SSS}. Consequently, dimensionality reduction is essential not only to engineering applications but also to the design of classifiers. In fact, the design of a classifier becomes extremely simple if all patterns of the same class hold the same feature vector while hold different feature vectors between classes.

Up to now, a large family of algorithms had been designed to provide different solutions to the problem of DR. Among them, the linear algorithms Principal Component Analysis (PCA) \cite{PCA} and Linear Discriminative Analysis (LDA) \cite{LDA} had been the two most popular methods due to their relative simplicity and effectiveness. However, PCA and LDA considered only the global scatter of training samples and they failed to reveal the essential data structures nonlinearly embedded in a high dimensional space. To overcome these limitations, the manifold learning methods were proposed by assuming that the data lie in a low dimensional manifold of the high dimensional space \cite{LLE}. Locality Preserving Projection (LPP) \cite{LPP} was one of the representative manifold learning methods. Success of manifold learning implies that the high dimensional facial images can be sparsely represented or coded by the representative samples on the manifold. Very recently, Wright \emph{et al.} presented a Sparse Representation based Classification (SRC) method for face recognition \cite{SRC}. The main idea of SRC is to represent a given test sample as a sparse linear combination of all training samples, the nonzero sparse representation coefficients are supposed to concentrate on the training samples with the same class label as the test sample. SRC shows that the classification performance of most meaningful features converges when the feature dimension increases if a SRC classifier is used. Although this does provide some new insights into the role of feature extraction played in a pattern classification tasks, Qiao \emph{et al.} \cite{SPP} argued that designing an effective and efficient feature extractor is still of great importance since the classification algorithm could become simple and tractable, and a unsupervised DR method called Sparsity Preserving Projections (SPP) was proposed, which aimed to preserve the sparse reconstructive relationship of the data in low-dimensional subspace. Yang and Chu \cite{SRCDP} proposed a Sparse Representation Classifier steered Discriminative Projection (SRC-DP) method. It used the decision rule of SRC to steer the design of a dimensionality reduction method. SRC-DP iteratively obtained the projection matrix and spare coding coefficient of each training data. But the convergence of SRC-DP was not clear, and also it was time consuming due to the large computing cost of iterative sparse coding.

In this paper, to enhance the recognition performance of SR, we propose a supervised DR method base on sparse representation, which is named the Optimized Projection for Sparse Representation based Classification (OP-SRC). Similar to SRC-DP, OP-SRC aims to gain a discriminative projection such that SRC achieves the optimum performance in the transformed low-dimensional space. Since SRC predicts the class label of a given test sample based on the representational residual, OP-SRC utilizes the label information to enhance the residuals more informative. We will also show that OP-SRC is naturally orthogonal, which may help preserve the shape of the data distribution.

The remainder of this paper is organized as follows: Section 2 reviews the SRC algorithm. Section 3 presents the OP-SRC method. The experimental results are presented in Section 4 and some discussions will be presented based on the results on several databases. Finally, we conclude this paper in Section 5.
\section{Sparse Representation based Classification}
Given sufficient $c$ classes training samples, a basic problem in pattern recognition is to correctly determine the class which a new coming (test) sample belongs to. We arrange the $n_i$ training samples from the $i$-th class as columns of a matrix $X_i=[x_{i1},\cdots,x_{in_i}]\in\mathbb{R}^{m\times n}$, where $m$ is the dimension. Then we obtain the training sample matrix $X=[X_1,\cdots,X_c]$,  where $n=\sum_{i=1}^{c}n_i$ is the total number of training samples.

Under the assumption of linear representation, a test sample $y\in\mathbb{R}^m$ will approximately lie on the linear subspace spanned by training samples
\begin{equation}\label{linrep}
    y=X\alpha\in\mathbb{R}^m
\end{equation}
If $m<n$, the system of Eq. (\ref{linrep}) is underdetermined, and also, its solution is not unique. This motivates us to seek the sparest solution to Eq. (\ref{linrep}), by solving the following $\ell^0$-minimization problem:
\begin{equation}\label{L0min}
    (\ell^0): \  \hat{\alpha}_0=\arg\min ||\alpha||_0 \ \ \mbox{subject to} \ \ y=X\alpha,
\end{equation}
where $||\cdot||_0$ denotes the $\ell^0$-norm, which counts the number of nonzero entries in a vector. However, the problem of finding the sparsest solution of an underdetermined system of linear equations is NP-hard and difficult even to approximate \cite{L0L1}.
The theory of compressive sensing \cite{L0L1equ} \cite{L0L1equ2} reveals that if the solution to the $\ell^0$-minimization problem is sparse enough, then it is equal to the following $\ell^1$-minimization problem
\begin{equation}\label{L1min}
    (\ell^1): \ \hat{\alpha}_1=\arg\min ||\alpha||_1 \ \ \mbox{subject to} \ \ y=X\alpha,
\end{equation}
In order to deal with occlusion, the $\ell^1$-minimization problem is extended to the stable $\ell^1$-minimization problem as follow:
\begin{equation}\label{sL1min}
    (\ell^1_s): \ \hat{\alpha}_1=\arg\min ||\alpha||_1 \ \ \mbox{subject to} \ \ ||y-X\alpha||_2\leq\varepsilon,
\end{equation}
where $\varepsilon$ is a given tolerance.

For a given test sample $y$, SRC first computes its sparse representation coefficient $\hat{\alpha}_1$ by solving the $\ell^1$-minimization problem (\ref{L1min}) or (\ref{sL1min}), then determines the class of this test sample from its reconstruction error between this test sample and the training samples of class $i$,
\begin{equation}\label{resi_comp}
    r_i(\alpha)=||y-X\delta_i(\alpha)||_2.
\end{equation}
For each class $i$, $\delta_i(\alpha):\mathbb{R}^n\rightarrow\mathbb{R}^n$ is the characteristic function which selects the coefficients associated which the $i$-th class. Then the class $C(y)$ which the test sample $y$ belongs to is determined by
\begin{equation}\label{resi_deci}
    C(y)=\arg\min_i r_i(\alpha).
\end{equation}
	
SRC is robust to noise and performs well for face recognition, it attracts much attention in recent years and boosts the research of sparsity based machine learning. Elhamifa and Vidal \cite{StructuredSR} proposed a more robust classification method using structured sparse representation, while Gao \emph{et al.} \cite{KSRC} introduced a kernel version of SRC. In \cite{L1Graph}, the $\ell^1$-graph was established by sparsely coding one sample over the other samples for clustering. But in this paper, we focus on the sparse representation based dimensionality reduction problem, not the extension of SRC. A discriminative learning method is presented in the next section.
\section{Optimized Projection for Sparse Representation based Classification}
In this section, we consider the supervised DR problem. Considering a training sample $x$ (belonging to the $i$-th class) and its sparse representation coefficient $\alpha$ based on other training samples as a dictionary. Ideally, the entries of $\alpha$ are zero except those associated with the $i$-th class. In many practical face recognition scenarios, the training sample $x$ could be partially corrupted or occluded. Or sometimes the training samples are not enough to represent the given sample. In these cases, the residual associated with the $i$-th class $r_i(x)$ may be not small enough, and may produce an erroneous predict. Thus, the Optimized Projection for Sparse Representation based Classification (OP-SRC) is proposed which aims to seek a linear projection matrix such that in the transformed low-dimensional space, the within-class reconstruction residual is as small as possible and simultaneously the between-class reconstruction residual is as large as possible.

\begin{figure*}[!t]
\centering
\includegraphics[width=1\textwidth]{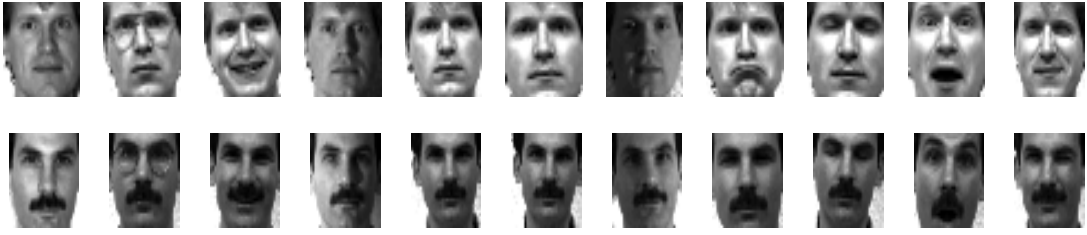}
\caption{Samples of two subjects from the Yale database.}
\label{Fig_yale}
\end{figure*}
\begin{table*}[!t]
\centering
\caption{Mean recognition rates (\%) and standard deviations on the Yale database.}
\label{Tab_resYale}
\centering
\begin{tabular}{c c c c c}
\hline
Methods &   4 Train             &   5 Train             &   6 Train             &    7 Train            \\\hline
PCA &   0.6467¡À0.044(52)       &   0.671¡À0.029(64)    &   0.725¡À0.042(88)    &   0.721¡À0.055(64)    \\
LDA     &   0.717¡À0.057(14)    &   0.752¡À0.039(14)    &   0.799¡À0.046(14)    &   0.814¡À0.046(14)    \\
SPP     &   0.607¡À0.049(57)    &   0.638¡À0.045(72)    &   0.676¡À0.044(88)    &   0.702¡À0.046(104)   \\
SRC-DP  &   0.706¡À0.049(29)    &   0.724¡À0.035(37)    &   0.771¡À0.042(34)    &   0.773¡À0.043(43)    \\
OP-SRC  &   \textbf{0.758¡À0.048(48)}    &   \textbf{0.794¡À0.036(62)}    &   \textbf{0.833¡À0.038(74)}    &   \textbf{0.853¡À0.048(88)}    \\\hline
\end{tabular}
\end{table*}

Let $P\in\mathbb{R}^{m\times d}$  be the optimized projection matrix with $d\ll m$. The data matrix in the original input space $\mathbb{R}^m$ are mapped into a $d$-dimensional space $\mathbb{R}^d$, that is, $Y=P^TX$. For each training sample $y_{ij}=P^Tx_{ij}$ from $Y$ in the transformed $d$-dimensional space $\mathbb{R}^d$, by solving the extended $\ell^1$-minimization problem (\ref{sL1min}), we obtain its sparse coding coefficient $\alpha_{ij}$ by using the remaining training samples as a dictionary. Based on the decision rule of SRC, we define the within-class residual matrix as follows
\begin{equation}\label{Rw_Y}
    \tilde{R}_W=\frac{1}{n}\sum_{i=1}^{c}\sum_{j=1}^{n_i}(y_{ij}-Y\delta_i(\alpha_{ij}))(y_{ij}-Y\delta_i(\alpha_{ij}))^T.
\end{equation}
The between-class residual matrix is defined as follow
\begin{equation}\label{Rb_Y}
    \tilde{R}_B=\frac{1}{n(c-1)}\sum_{i=1}^{c}\sum_{j=1}^{n_i}\sum_{l\neq i}(y_{ij}-Y\delta_l (\alpha_{ij}))(y_{ij}-Y\delta_l (\alpha_{ij}))^T.
\end{equation}
The total residual matrix is defined as follow
\begin{eqnarray}\label{Rt_y}
    \tilde{R}_T&=&\frac{n\tilde{R}_W+n(c-1)\tilde{R}_B}{nc} \\
                &=&\frac{1}{nc}\sum_{i=1}^c\sum_{j=1}^{n_i}\sum_{l=1}^c (y_{ij}-Y\delta_l(\alpha_{ij}))(y_{ij}-Y\delta_l(\alpha_{ij}))^T.
\end{eqnarray}

To make SRC perform well on training data, we expect that the within-class residual is as small as possible and simultaneously the between-class residual is as large as possible. Therefore, we can choose to maximize the following criterion \cite{MMC}
\begin{equation}\label{criterion}
    J(P)=tr(\beta\tilde{R}_B-\tilde{R}_W),
\end{equation}
where $\beta$ is the weight parameter which balances the between-class and within-class residual information. Since $P$ is a linear mapping, it is easy to show $\tilde{R}_W=P^TR_WP$ and $\tilde{R}_B=P^TR_BP$, where
\begin{equation}\label{R_w}
    R_W=\frac{1}{n}\sum_{i=1}^c\sum_{j=1}^{n_i}(x_{ij}-X\delta_i(\alpha_{ij}))(x_{ij}-X\delta_i(\alpha_{ij}))^T,
\end{equation}
\begin{equation}\label{R_b}
\raggedleft
    R_B=\frac{1}{n(c-1)}\sum_{i=1}^c\sum_{j=1}^{n_i}\sum_{l\neq i} (x_{ij}-X\delta_l(\alpha_{ij}))(x_{ij}-X\delta_l(\alpha_{ij}))^T.
\end{equation}
So, we have
\begin{equation}\label{criterion2}
    J(P)=tr(P^T(\beta R_B-R_W)P).
\end{equation}

In order to avoid degenerate solutions, we additionally require that $P$ is constituted by the unit vectors, \emph{i.e.} $P=[p_1,\cdots,p_d]$ and $p_k^Tp_k=1$, $k=1,\cdots,d$. One may use other constraints. For example, we can require $tr(P^TR_WP)=1$ and then maximize $tr(P^TR_BP)$. The motivation by using the constraint $p_k^Tp_k=1$ is that it will result to an orthogonal projection, which may help preserve the shape of the data distribution. Thus, the objective function can be recast as the following optimization problem:
\begin{eqnarray}\label{maximipro}
   & & \max  \sum_{k=1}^d p_k^T(\beta R_B-R_W)p_k \\\nonumber
   & & \mbox  {subject to} \ p_k^Tp_k=1,k=1,\cdots,d.
\end{eqnarray}
We can use the Lagrange multipliers to transform the above objective function to include the constraint
\begin{equation}\label{Lagrange}
    L(p_k,\lambda_k)=\sum_{k=1}^d p_k^T(\beta R_B-R_W)-\lambda_k(p_k^Tp_k-1).
\end{equation}
The optimization is performed by setting the partial derivative of $L$ with respect to $p_k$ to zero
\begin{equation}\label{Larg}
    \frac{\partial L}{\partial p_k}=(\beta R_B-R_W-\lambda_k I)p_k=0, k=1,\cdots,d.
\end{equation}
Now we obtain
\begin{equation}\label{Larg2}
	 	(\beta R_B-R_W)p_k=\lambda_k p_k, k = 1,\cdots,d,
\end{equation}
which means that the $\lambda_k$'s are the eigenvalues of $\beta R_B-R_W$ and the $p_k$'s are the corresponding eigenvectors. Thus
\begin{equation}\label{JP}
J(P)=\sum_{k=1}^d p_k^T(\beta R_B-R_W)p_k=\sum_{k=1}^d\lambda_k p_k^Tp_k=\sum_{k=1}^d\lambda_k.
\end{equation}
Therefore, $P$ is composed of the first $d$ largest eigenvectors of $\beta R_B-R_W$ and $J(P)$ is maximized.

The solution of the optimization problem (\ref{maximipro}) has the following property:

\newtheorem{thm}{Theorem}
\newtheorem{lem}[thm]{Proposition}
\begin{lem}
The columns of the optimal solution $P$ to the optimization problem (\ref{maximipro}) are orthogonal, that is, $p_i^Tp_j=0$, for any $i\neq j$, and $p_i^Tp_i=1$.
\end{lem}
It is easy to prove the orthogonality of solution $P$ due to the symmetry of $(\beta R_B-R_W)$. Thus, OP-SRC is an supervised orthogonal projection method which may preserve more discriminative information for classification, expecially for the SRC method.

\begin{figure*}[!t]
\centering
\includegraphics[width=1\textwidth]{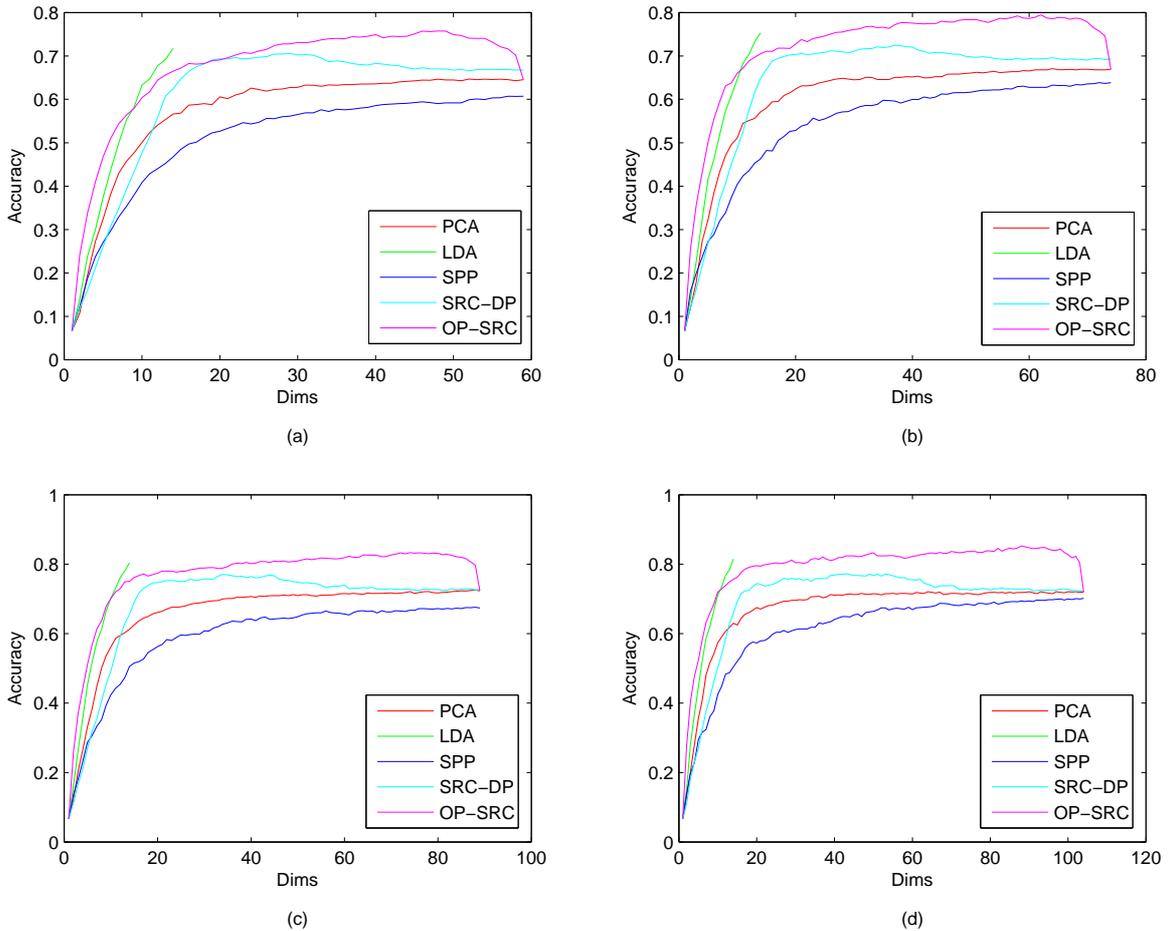}
\caption{Accuracy rates versus reduced dimensions on the Yale database: (a) 4 Train; (b) 5 Train; (c) 6 Train; (d) 7 Train.}
\label{Fig_res_yale}
\end{figure*}

\section{Experimental Verification}
In this section, we investigate the performance of our proposed OP-SRC method for face representation and recognition. The system performance is compared with PCA \cite{PCA}, LDA \cite{LDA}, SPP \cite{SPP} and SRC-DP \cite{SRCDP}. PCA and LDA are two of the most popular linear methods in FR. SPP and SRC-DP are two new methods corresponding to sparse representation. Similar to SPP and SRC-DP, we first perform PCA to reduce the dimension before implementing OP-SRC. Finally, SRC is employed for classification.

\subsection{Data Sets and Experimental Settings}
We test our proposed method on three popular face databases, including Yale \cite{LDA}, ORL \cite{FSSamaria} and UMIST \cite{DBGraham}. There are wide-range variations, including pose,  illumination, and gesture alterations, existing in the databases. For these databases, we randomly select part of the images per class for training (\emph{i.e.} 4, 5, 6, and 7 of 11 images per subject for Yale, 4, 5, 6 and 7 of 10 images per subject for ORL and 6, 8, 10 and 12 of about 29 images per subject for UMIST), and the remainder for test. In particular, with the given training set, the projection $P$ is learned by PCA, LDA, SPP, SRC-DP and OP-SRC\footnote{The Matlab code can be found from our homepage: http://mail.ustc.edu.cn/~canyilu/}, respectively, and the test samples are subsequently transformed by the learned projection. Then specific classifier is employed to evaluate the recognition rates on the test data, and SRC is used in this paper.

In the experiments, the images are cropped to a size of $32\times 32$, and the gray level values of all images are rescaled to [0,1]. 20 training/test splits are randomly generated and the average classification accuracies over these splits are reported in tables and figures.

The SPAMS package \cite{SPAMS1} \cite{SPAMS2} is used for solving the extended $\ell^1$-minimization problem (4). In our experiments, we experimentally set $\varepsilon=0.05$ (refer to (4)) which usually leads SRC to better performance than other parameters, and set $\beta=0.25$ (refer to (14)) by searching in a large range of candidates.
\begin{figure*}[!t]
\centering
\includegraphics[width=1\textwidth]{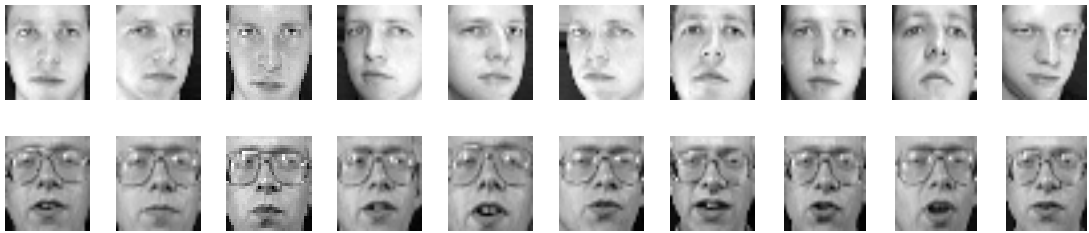}
\caption{Samples of two subjects from the ORL database.}
\label{Fig_ORL}
\end{figure*}
\begin{table*}[!t]
\centering
\caption{Mean recognition rates (\%) and standard deviations on the ORL database.}
\label{Tab_resORL}
\centering
\begin{tabular}{c c c c c}
\hline
Methods &   4 Train             &   5 Train             &   6 Train             &    7 Train            \\\hline
PCA     &   0.898¡À0.019(127)   &   0.921¡À0.018(183)   &   0.941¡À0.018(193)   &   0.954¡À0.023(134)   \\
LDA     &   0.899¡À0.019(39)    &   0.930¡À0.017(39)    &   0.941¡À0.019(39)    &   0.950¡À0.023(39)    \\
SPP     &   0.861¡À0.018(108)   &   0.887¡À0.026(170)   &   0.903¡À0.031(180)   &   0.922¡À0.026(202)   \\
SRC-DP  &   0.888¡À0.018(124)   &   0.918¡À0.018(131)   &   0.929¡À0.028(221)   &   0.943¡À0.022(190)   \\
OP-SRC  & \textbf{0.925¡À0.017(153)}	& 	\textbf{0.950¡À0.017(195)}		& \textbf{0.968¡À0.015(224)}	& 	\textbf{0.975¡À0.013(255)}
\\\hline
\end{tabular}
\end{table*}
\subsection{Yale Database}
The Yale database contains 165 gray scale images of 15 individuals. It was constructed at the Yale Center for Computational Vision and Control. The images demonstrate variations in lighting condition, facial expression (normal, happy, sad, sleepy, surprised, and wink). Figure \ref{Fig_yale} shows some samples of two subjects of the Yale database. A random subset with $l$ (=4, 5, 6, 7) images per individual is taken with labels to form the training set, and the rest of the database is considered to be the test set. For each given $l$, we average the recognition accuracy over 20 random splits. Notice that LDA is different from other methods because the maximal number of dimension is less than the number of class $c$ \cite{LDA}.

In general, the performance of all these methods varies with the number of dimensions. We show the best results and the optimal dimensions obtained by PCA, LDA, SPP, SRC-DP and OP-SRC in Table \ref{Tab_resYale}, including the mean of accuracies as well as the standard deviations.

From Table \ref{Tab_resYale}, it can be found that OP-SRC obtains the highest recognition rates in all cases. Figure \ref{Fig_res_yale} shows the plots of accuracy rates versus reduced dimensions. Note that, when the dimension of feature continues to increase, the performance of the OP-SRC algorithm decreases and has the same accuracy with PCA on the highest dimension. In this case, the obtained optimized projection matrix $P$ is square and orthogonal, that is $P^TP=PP^T=I$. Thus, $||P^Tx-P^TX\alpha||_2=||x-X\alpha||_2$. The sparse representation coefficient in the transformed space will be the same as in the subspace projected by PCA. Thus, they always obtain the same recognition result.

\begin{figure*}[!t]
\centering
\includegraphics[width=1\textwidth]{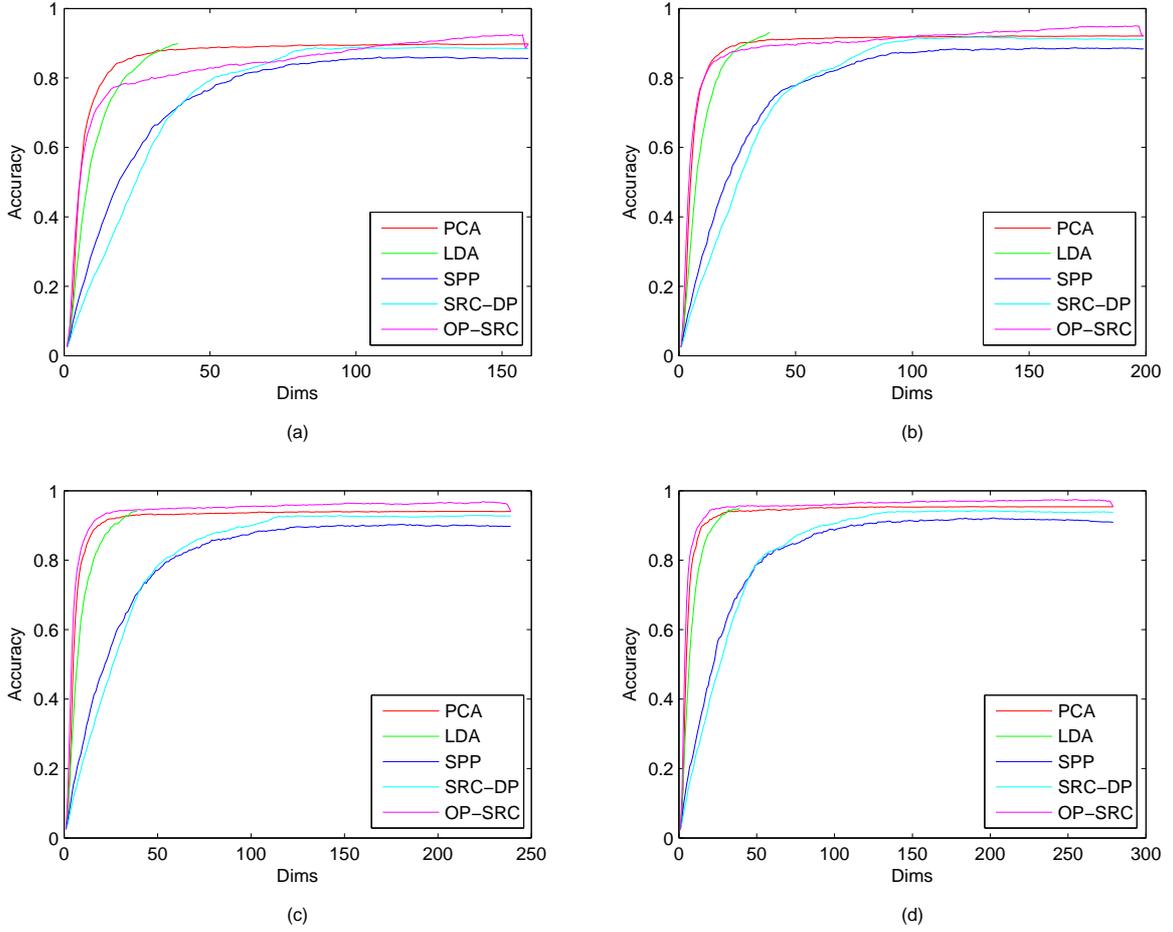}
\caption{Accuracy rates versus reduced dimensions on the ORL database: (a) 4 Train; (b) 5 Train; (c) 6 Train; (d) 7 Train.}
\label{Fig_res_ORL}
\end{figure*}
\subsection{ORL Database}
The ORL database consists of 10 face images from 40 subjects for a total of 400 images, with some variations in poses, facial expressions and details. Some images were captured at different times and had different variations including expression (open or closed eyes, smiling or nonsmiling) and facial details (glasses or no glasses). The images were taken with a tolerance for some tilting and rotation of the face up to 20 degrees. Figure 3 shows some samples of two subjects of the ORL database. A random subset with $l$ (=4, 5, 6, 7) images per individual is taken with label to form the training set. The rest of the database is considered to be the test set. The experimental protocol is the same as that on the Yale database. The recognition results are shown in Table 2 and Figure 4.

From Table \ref{Tab_resORL} and Figure \ref{Fig_res_ORL}, we find that most dimensionality reduction methods perform well, since the variation of faces in the ORL database is limited. PCA is even more accurate than SRC-DP which is supervised. If the number of training data is small, \emph{i.e.} 4 and 5 samples of each subject for training, OP-SRC also performs worse than PCA in low-dimensional space, but much better in high-dimensional space. It seems that OP-SRC may lead to overfitting on the ORL database in low-dimensional space with limited training data.
\subsection{UMIST Database}
\begin{figure*}[!t]
\centering
\includegraphics[width=1\textwidth]{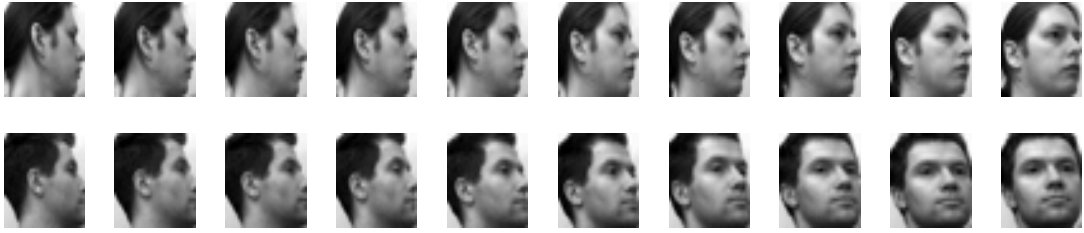}
\caption{Samples of two subjects from the UMIST database.}
\label{Fig_UMIST}
\end{figure*}

The UMIST database contains 564 images of 20 individuals, each covering a range of poses from profile to frontal views. The subjects cover a range of race, sex and appearance. We use a cropped version of the UMIST database that is publicly available at S. Roweis' web page\footnote{http://cs.nyu.edu/~roweis/data.html}. Figure \ref{Fig_UMIST} shows some images of two subjects of the UMIST database. We randomly select $l$ (=6, 8, 10, 12) images from each individual for training, and the rest for test.

\begin{figure*}[!t]
\centering
\includegraphics[width=1\textwidth]{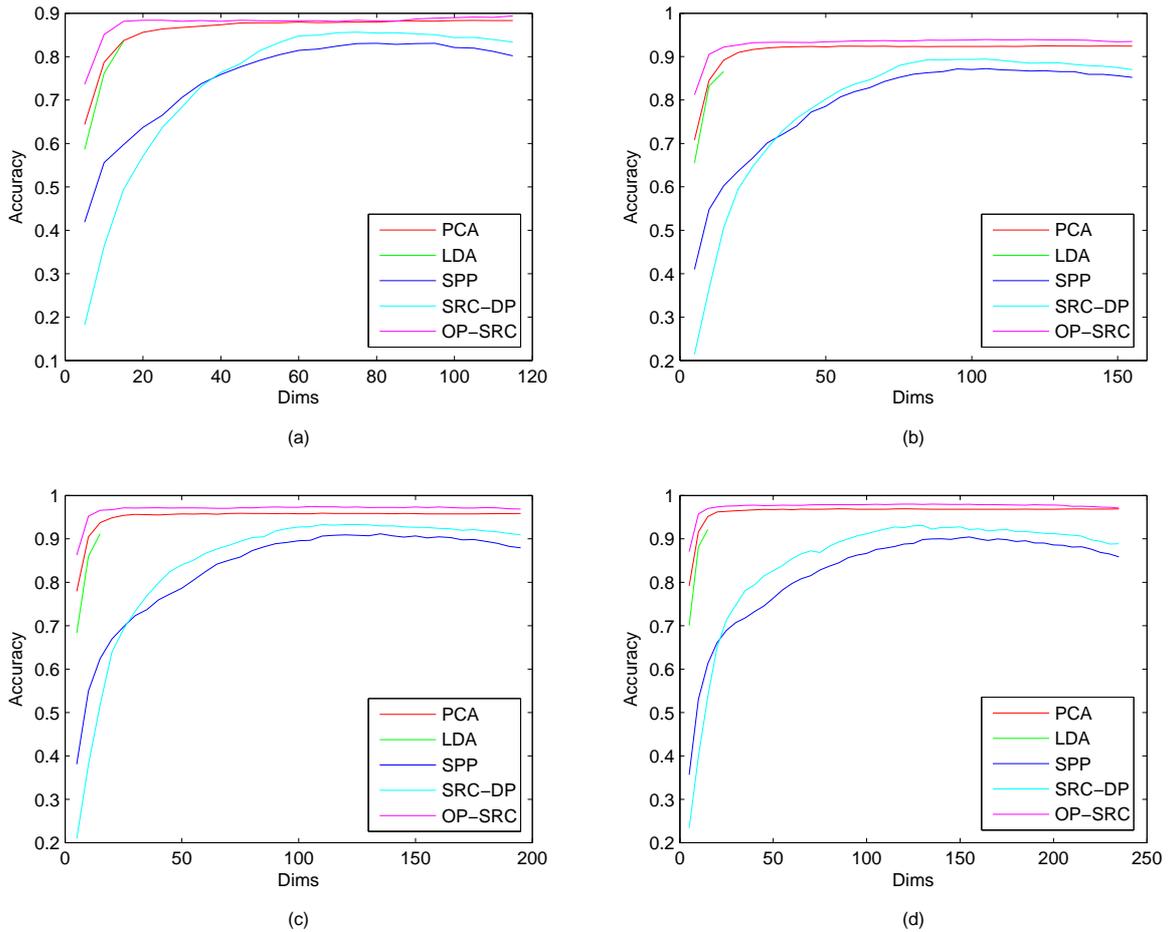}
\caption{Accuracy rates versus reduced dimensions on the UMIST database: (a) 6 Train; (b) 8 Train; (c) 10 Train; (d) 12 Train.}
\label{Fig_resUMIST}
\end{figure*}

Table \ref{Tab_resUMIST} gives the best classification accuracy rates and the corresponding standard deviations of five algorithms under different sizes of the training set. Figure \ref{Fig_resUMIST} plots the recognition rates of five algorithms under different reduced dimensions when the size of training samples from each class is 6, 8, 10 and 12, respectively. From Table \ref{Tab_resUMIST} and Figure \ref{Fig_resUMIST}, we find that OP-SRC outperforms the other methods in different dimensions and different numbers of training data setting.
\subsection{Discussions}
Based on the results on the Yale, ORL and UMIST databases, we draw the following observations and discussions:
\begin{enumerate}
\item OP-SRC always outperforms PCA, SPP and SRC-DP on the Yale and UMIST databases, and also is more accurate than PCA when the subspace dimension exceeds a certain threshold on the ORL database. OP-SRC even performs better than LDA in low-dimensional subspace on the ORL and UMIST databases. The top average recognition rates of OP-SRC are much higher than PCA, SPP, LDA and SRC-DP on these three databases. The superior of OP-SRC comes from its orthogonality and matching well with the SRC algorithm.
\item Similar to other dimensionality reduction methods, the recognition accuracy of OP-SRC will first increase according to the dimensions, but decrease at last and obtain the same result as PCA on the highest dimension. This is because the data is first projected onto a PCA subspace, and the $\ell^2$-norm is invariant to orthogonal the OP-SRC projection on the highest dimension.
\item From our experiments, we also find that OP-SRC is more efficient than SPP and SRC-DP which are all spare coding based methods. It is more practical for real applications.
\end{enumerate}
\begin{table*}[!t]
\centering
\caption{Mean recognition rates (\%) and standard deviations on the UMIST database.}
\label{Tab_resUMIST}
\centering
\begin{tabular}{c c c c c}
\hline
Methods &   6 Train             &   8 Train             &   10 Train            &    12 Train            \\\hline
PCA     &   88.35¡À2.32(105)    &   92.48¡À3.13(125)    &   0.9592¡À1.29(110)   &   96.93¡À1.84(85)     \\
LDA     &   83.54¡À1.82(15)     &   86.58¡À3.17(15)     &   91.15¡À1.26(15)     &   92.18¡À1.68(15)     \\
SPP	    &   83.08¡À2.69(80)	    &   87.25¡À2.64(105)    &   91.17¡À2.13(135)    &   90.45¡À2.78(155)    \\
SRC-DP  &	85.63¡À2.20(75)     &   89.42¡À2.73(105)	&   93.28¡À1.50(120)	&   93.07¡À2.48(130)    \\
OP-SRC	&   \textbf{89.41¡À1.93 (115)}	&   \textbf{93.93¡À2.98(105)}	&   \textbf{0.9744¡À1.19(105)}	&   \textbf{98.00¡À1.57(120)}
\\\hline
\end{tabular}
\end{table*}
\section{Conclusions}
In this paper, based on sparse representation, we propose a new algorithm called Optimized Projection for Sparse Representation based Classification (OP-SRC) for supervised dimensionality reduction. The optimized projection of SRC decreases the within-class reconstruction residual and simultaneously increases the between-class reconstruction residual which matches with SRC optimally in theory. OP-SRC is orthogonal which may help preserve more discriminative information for classification. The experimental results on the three face databases clearly demonstrate that the proposed OP-SRC has much better performance than PCA, LDA, SPP and SRC-DP, and also it is more effective with respect to the sparse representation based classification.
\section*{Acknowledgment}
This work was supported by the grants of the National Science Foundation of China, Nos. 60975005, 61005010, 60873012, 60805021, 60905023, 31071168, 61133010, and the Knowledge Innovation Program of the Chinese Academy of Sciences, Y023A61121.
\bibliographystyle{elsarticle-num}
\bibliography{OPSRCref}

\parpic{\includegraphics[width=0.7in,height=1.1in,clip,keepaspectratio]{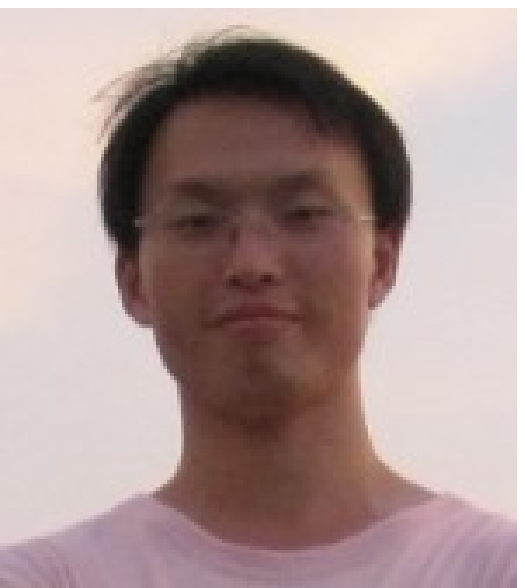}}
\noindent {\bf Can-Yi Lu} received the B.S. degree in Information and Computing Science \& Applied Mathematics from Fuzhou University (FZU), Fuzhou, China, in 2009. Now, he is a candidate of the master degree in Pattern Recognition \& Intelligent Systems from University of Science and Technology of China (USTC), Hefei, China. His research interests include sparse representation and low rank based machine learning and applications.

\parpic{\includegraphics[width=0.8in,height=1.1in,clip,keepaspectratio]{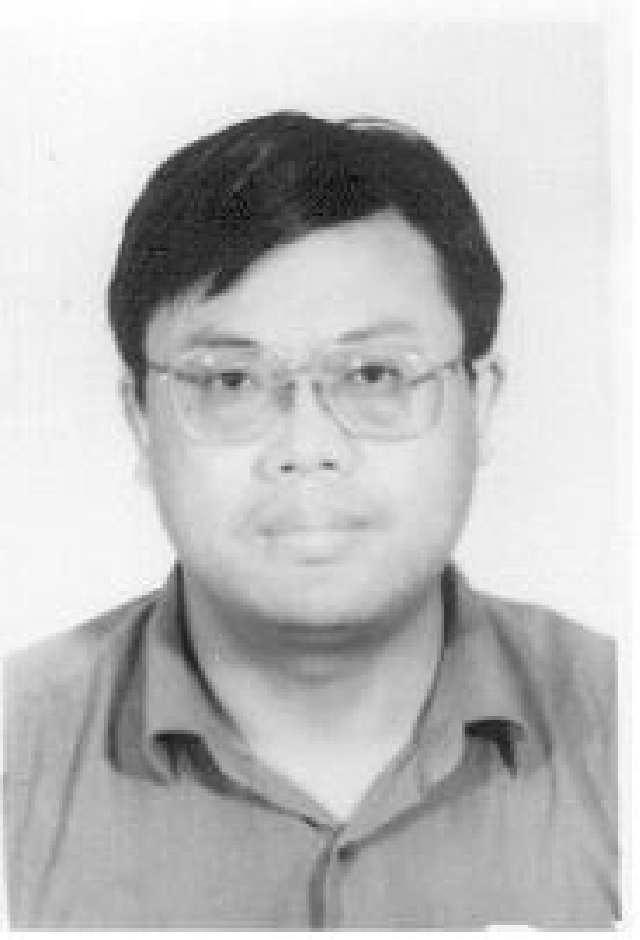}}
\noindent {\bf De-Shuang Huang} received the B.Sc., M.Sc. and Ph.D. degrees all in electronic engineering from Institute of Electronic Engineering, Hefei, China, National Defense University of Science and Technology, Changsha, China and Xidian University, Xian, China, in 1986, 1989 and 1993, respectively. During 1993-1997 period he was a postdoctoral student respectively in Beijing Institute of Technology and in National Key Laboratory of Pattern Recognition, Chinese Academy of Sciences, Beijing, China. In Sept, 2000, he joined the Institute of Intelligent Machines, Chinese Academy of Sciences as the Recipient of ¡°Hundred Talents Program of CAS¡±. In September 2011, he entered into Tongji University as Chaired Professor. From Sept 2000 to Mar 2001, he worked as Research Associate in Hong Kong Polytechnic University. From Aug. to Sept. 2003, he visited the George Washington University as visiting professor, Washington DC, USA. From July to Dec 2004, he worked as the University Fellow in Hong Kong Baptist University. From March, 2005 to March, 2006, he worked as Research Fellow in Chinese University of Hong Kong. From March to July, 2006, he worked as visiting professor in Queen¡¯s University of Belfast, UK. In 2007, 2008, 2009, he worked as visiting professor in Inha University, Korea, respectively. At present, he is the head of Machines Learning and Systems Biology Laboratory, Tongji University.

Dr. Huang is currently a Senior member of the IEEE. He has published over 200 papers. Also, in 1996, he published a book entitled ¡°Systematic Theory of Neural Networks for Pattern Recognition¡± (in Chinese), which won the Second-Class Prize of the 8th Excellent High Technology Books of China, and in 2001 \& 2009 another two books entitled ¡°Intelligent Signal Processing Technique for High Resolution Radars¡± (in Chinese) and ¡°The Study of Data Mining Methods for Gene Expression Profiles¡± (in Chinese), respectively. In addition, he was the PhD advisor in the University of Science and Technology of China. His current research interest includes bioinformatics, pattern recognition and machine learning.
\end{document}